\documentclass[conference]{IEEEtran}
\IEEEoverridecommandlockouts
\usepackage{cite}
\usepackage{amsmath,amssymb,amsfonts}
\usepackage{algorithmic}
\usepackage{graphicx}
\usepackage{textcomp}
\usepackage{xcolor}
\usepackage{booktabs} %
\usepackage{hyperref}
\usepackage{fontawesome5}

\def\BibTeX{{\rm B\kern-.05em{\sc i\kern-.025em b}\kern-.08em
    T\kern-.1667em\lower.7ex\hbox{E}\kern-.125emX}}
\begin{document}

\title{Towards a Unified Representation Evaluation Framework Beyond Downstream Tasks
\thanks{Christos Plachouras and Julien Guinot are supported by UK Research and Innovation (grant number EP/S022694/1).}
}

\author{
\IEEEauthorblockN{
Christos Plachouras$^{1}$,
Julien Guinot$^{1,2}$,
György Fazekas$^{1}$,
Elio Quinton$^{2}$,
Emmanouil Benetos$^{1}$,
Johan Pauwels$^{1}$
}
\IEEEauthorblockA{
$^{1}$School of Electronic Engineering and Computer Science, Queen Mary University of London, London, UK\\
$^{2}$Music \& Audio Machine Learning Lab, Universal Music Group, London, UK}
\{c.plachouras, j.guinot, g.fazekas, emmanouil.benetos, j.pauwels\}@qmul.ac.uk, elio.quinton@umusic.com
}

\maketitle

\begin{abstract}
Downstream probing has been the dominant method for evaluating model representations, an important process given the increasing prominence of self-supervised learning and foundation models. However, downstream probing primarily assesses the availability of task-relevant information in the model's latent space, overlooking attributes such as equivariance, invariance, and disentanglement, which contribute to the interpretability, adaptability, and utility of representations in real-world applications. While some attempts have been made to measure these qualities in representations, no unified evaluation framework with modular, generalizable, and interpretable metrics exists.

In this paper, we argue for the importance of representation evaluation beyond downstream probing. We introduce a standardized protocol to quantify informativeness, equivariance, invariance, and disentanglement of factors of variation in model representations. We use it to evaluate representations from a variety of models in the image and speech domains using different architectures and pretraining approaches on identified controllable factors of variation. We find that representations from models with similar downstream performance can behave substantially differently with regard to these attributes. This hints that the respective mechanisms underlying their downstream performance are functionally different, prompting new research directions to understand and improve representations. 
\end{abstract}

\begin{IEEEkeywords}
representation evaluation, representation learning, equivariance, invariance, disentanglement
\end{IEEEkeywords}

\section{Introduction}

Representation learning has become popular across many fields due to its effectiveness, computational efficiency, and the relative simplicity of using representations from pretrained models as features for various downstream tasks. Many architectures, training paradigms, and modalities have been used to learn representations that are effective in a variety of tasks, such as retrieval, classification, and generation.

Downstream evaluation, typically using shallow probing, provides some insight into the suitability of a representation for a target downstream task. In the traditional representation evaluation framework \cite{grill2020bootstrap, huang2022audiomae}, shallow probes are trained on a range of downstream tasks using frozen representations extracted from pretrained models. In this way, the availability of task-relevant information within the pretrained latent space is evaluated, such as the linear or non-linear separability of classes for classification tasks. However, downstream evaluation does not directly reveal the intrinsic structure and organization of the latent space, which can be equally relevant for tasks like retrieval and generation.

While there exists prior work focusing specifically on these attributes of representations learned from self-supervision, there is, to the best of our knowledge, no standardized protocol or methodological grounding for evaluating representations beyond downstream tasks. We argue that the field misses a more global perspective on representation evaluation by focusing almost entirely on downstream performance and that evaluating representations through other axes can benefit the community by uncovering the structural impact of different pretraining approaches on learned representations.

In this work, we propose an extended representation evaluation framework organizing representation quality into several quantified and formalized axes: informativeness, equivariance, invariance, and disentanglement. We argue for the importance of these attributes for specific applications, and we introduce a standardized protocol for evaluating representation models on these axes, intending to show that models that exhibit similar performance on downstream tasks can have drastically different underlying mechanisms in other desirable attributes for representations. We demonstrate this through experiments in the image and speech domains, evaluating a wide range of models with varying architectures and pretraining paradigms, and release \texttt{synesis},\footnote{\texttt{\url{https://github.com/chrispla/synesis}}} a software package for holistic representation evaluation following our framework.

\section{Towards holistic representation evaluation}\label{sec:position_and_conceptual_framing}

\subsection{Position}
\subsubsection{Downstream probing provides an incomplete view of representations}

Recent advances in SSL have been largely driven by the goal of achieving higher scores on increasingly extensive downstream evaluation benchmarks. This focus has led to significant progress, but it has overlooked other properties of representations, such as equivariance, invariance, and disentanglement. While downstream probing may indirectly engage these properties, directly evaluating them can provide more direct insight into the structural properties of representations. These insights are crucial for understanding the impact of different pretraining approaches and improving them, diagnosing representation weaknesses, selecting the most appropriate embeddings for specific applications, and improving the explainability, interactivity, and practical utility of pretrained models \cite{garridoSelfsupervisedLearningSplit2023, devillers2023equimod, xiao2021what}.

The limitations of downstream evaluation become particularly apparent when considering the ``Platonic representation hypothesis'' \cite{huh2024platonic}, which suggests that as model scale and training data increase, representations from different models converge. This seems counterintuitive when considering the fundamental mechanistic differences between approaches. For instance, contrastive methods explicitly learn invariances to certain transformations \cite{xiao2021what}, while generative approaches preserve fine-grained details for reconstruction. Structurally, joint embedding methods typically create hyperspherical embeddings through normalization, while Variational Autoencoders (VAEs) impose Gaussian priors on their latent space. These architectural and objective-driven differences suggest that even if performance converges at scale, the internal organization of information in the latent space likely remains distinct, and downstream evaluation does not reveal this.

\subsubsection{Standardized evaluation needs to mature}

The evaluation of learned representations has not always been as comprehensive as it is today in downstream tasks. Modern evaluation frameworks assess representation quality across multiple standardized datasets and tasks, such as analysis, segmentation, retrieval, and generation. At their core, these evaluations measure an implicit understanding of various ``factors of variation'' (FV) within the data. These factors can be broad (such as hundreds of classes in image classification or speaker identities in diarization), but they are ultimately composed of more granular factors, themselves decomposable. The ontology of these finer FVs defines the structure of the coarser ones. Over time, the standardization of tasks, datasets, and annotations for these factors has improved, providing a broader understanding of domain adaptability and downstream performance of representations.

No similarly mature framework exists for evaluating representations outside of downstream probing. We argue that advancing evaluation protocols in this direction is essential for the field of representation learning. We highlight emerging directions in representation evaluation and propose a standardized methodology for assessing new axes with given FVs. Similar to early efforts in downstream probing that focused on granular factors \cite{cifar, deng2012mnist}, our proposed evaluation also begins with foundational ``toy'' examples from image and speech representation learning. This intentional choice allows us to validate our approach while making it easier to keep the framework FV-agnostic.

The factors of variation we introduce here are not meant to be exhaustive. Instead, we emphasize the methodology of evaluating along our proposed axes (informativeness, equivariance, invariance, and disentanglement). Beyond simple transformations related to these factors, we encourage the community to develop new datasets and annotations for more complex transformations. Ultimately, this paper is not about prescribing which factors of variation to evaluate, but rather about establishing a structured approach to evaluating them and moving beyond downstream probing toward a more mature and standardized framework.

\subsection{Integrating evaluation axes}

Towards a unified, standardized, and formalized representation evaluation framework, we propose four additional axes of representation evaluation: \emph{informativeness}, \emph{equivariance}, \emph{invariance}, and \emph{disentanglement}.

\subsubsection{Informativeness}

We call informativeness the absolute information contained about a factor of variation within representations, i.e., how easy it is to predict a factor of variation from extracted representations. In this sense, informativeness is analogous to downstream probing, though the latter is often also used to describe the use of representations for an application, rather than purely for inspecting the content of representations. We keep informativeness as an axis of evaluation as it allows us to quantify the availability of information related to factors of variation in relation to other axes, and it anchors our understanding of representation behavior (given that the mechanistic interpretations of downstream probing are well explored) to better interpret representation behaviors for other axes.

\subsubsection{Equivariance}

Equivariance is a property of learned representations (and models by proxy) that describes the correspondence between transformations in the input space and their effects in the latent space. Formally, an encoder \(E\) is said to be equivariant with respect to a transformation \(\mathcal{T}\) if there exists $\mathcal{T'}$ such that:
\[
E(\mathcal{T}(x)) = \mathcal{T}'(E(x)),
\]
where \(\mathcal{T}'\) represents an equivalent transformation in the latent space. This property ensures that the structure of the transformation is preserved, allowing the model to retain information about \(\mathcal{T}\), in contrast to \(\mathcal{T}\)-invariance, where such information is discarded. Generally, \(\mathcal{T}\)-equivariance refers to the capacity of a model to maintain information about \(\mathcal{T}\) in its output, whereas a \(\mathcal{T}\)-invariant model discards this information. Quantifying equivariance in representations can help understand the behavior of models under transformations and knowingly apply these behaviors to downstream applications.

\subsubsection{Invariance}

Invariance reflects the stability of representations under perturbations or distribution shifts. It can be formalized by quantifying the change in representations when the input undergoes a modification. Given a perturbation \(\mathcal{T}\), the invariance of a representation can be evaluated as:
\[
\Delta_{\text{Invariance}} = \| z - \mathcal{E}(\mathcal{T}(x)) \|,
\]
where smaller values of \(\Delta_{\text{Invariance}}\) indicate greater stability under the applied perturbation. Invariance, i.e., the stability of representations under perturbation, is a specific case of \textit{robustness} of representations. 
Despite being extensively studied in the philosophy of science, where robustness analysis (RA) involves discerning invariance under multiple independent conditions, its definition in machine learning (ML) remains inconsistent and often vague \cite{freieslebenGeneralizationTheoryRobustness2023}. This prompts us to consider invariance as a barebones proxy for robustness that does not involve considerations about distribution robustness or generalizability (more on this in Section \ref{subsubsection: Other attributes}).  

\subsubsection{Disentanglement}\label{subsubsection: Disentanglement}

Disentanglement refers to a representation's ability to separate distinct and meaningful factors of variation so that changes in one latent dimension correspond to changes in a single interpretable concept while leaving others unaffected. In essence, disentangled representations ensure that modifying one factor of variation does not influence others within reasonable, distribution-realistic expectations. 

One commonly referred-to definition posits that a disentangled representation aligns each generative factor with a unique latent dimension, ensuring independence between non-related dimensions and FVs \cite{sepliarskaiaEvaluatingDisentangledRepresentations2019}. More formally, a representation \(z\) with latent dimensions \(z_i\) and factors of variation \(F_k\) can be described as disentangled if:
\[
I(z_i, z_{\neg i}) = 0 \quad \text{and} \quad I(z_i, F_k) = H(z_i, F_k),
\]
where \(I\) denotes mutual information, \(H\) is entropy, and \(z_{\neg i}\) represents all latent dimensions except \(z_i\). This implies that each latent dimension is independent of the others and fully informative about a corresponding generative factor.

\subsubsection{Other Attributes}\label{subsubsection: Other attributes}

Our selection of evaluation axes may seem restrictive given the broader set of proposed properties like robustness, interpretability, and generalizability \cite{zhang2022explainable,farahani2021brief,gilpin2018explaining}. We focus on these axes as core, structural attributes that can be evaluated independently of model architectures, training dynamics, or distributional shifts. Invariance, for instance, underpins robustness by measuring stability under controlled perturbations. Informativeness captures task-relevant information. Equivariance and disentanglement provide structural insights about applied perturbations that, among other things, can serve as building blocks for achieving interpretability and explainability. These concepts, however, are typically assessed at the model and prediction level, respectively, making their direct evaluation on a representation level difficult to define.

We consider generalizability best evaluated by expanding downstream benchmarks across diverse tasks and domains. Generalizability, in this context, emerges when representations maintain informativeness across new conditions, making task diversity a natural, scalable evaluation approach. Adaptability, while related, can be distinguished by its implication of model parameter updates to suit new tasks or data. This makes adaptability more a property of the model's learning capacity and architecture rather than an intrinsic characteristic of a static representation, thus also falling outside our direct focus on representation-level attributes.

Regarding robustness, we deliberately avoid its broader scope to isolate representation-specific stability. Freiesleben and Grote's theory of robustness in ML defines robustness via targets (e.g., representations) and modifiers (e.g., input perturbations) \cite{freieslebenGeneralizationTheoryRobustness2023}, with extrinsic factors like architectures, training protocols, and label distributions influencing outcomes. By focusing on invariance (representation stability under semantic-preserving transformations), we capture a fundamental component of robustness without conflating it with model-level generalization mechanisms or task-dependent factors.

\begin{figure*}[t]
    \centering
    \includegraphics[width=1\linewidth]{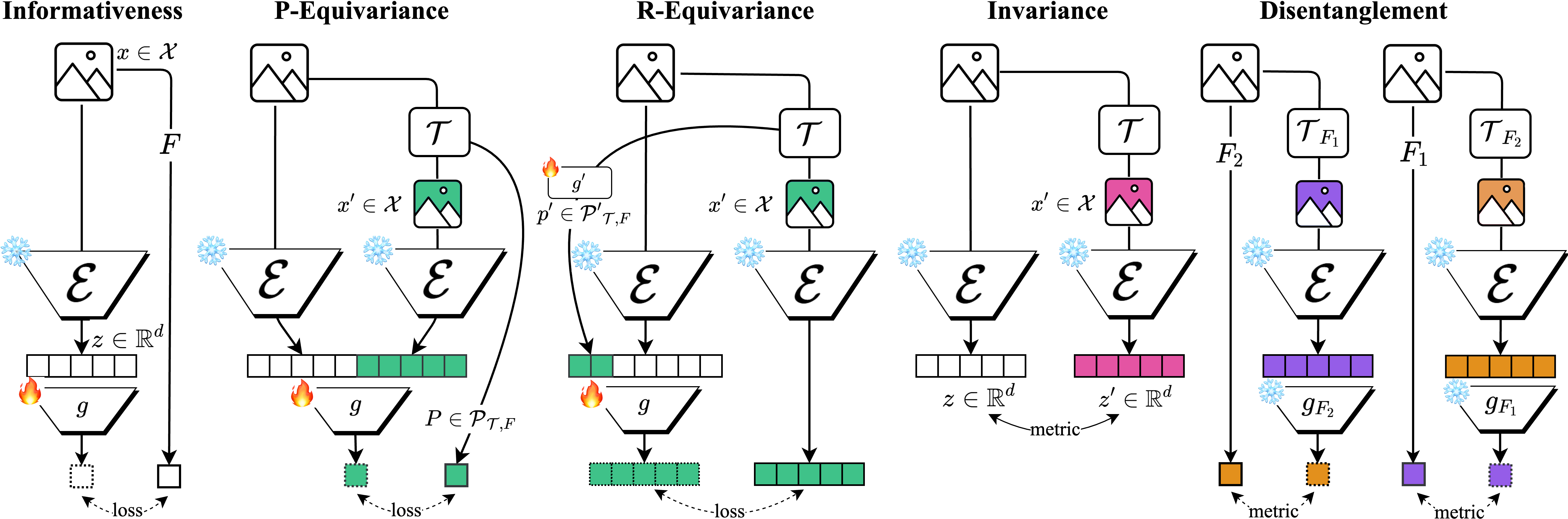}
    \caption{High-level overview of the evaluation approaches for informativeness, equivariance, invariance, and disentanglement. $\mathcal{E}$ are feature extractors, $g$ are projectors. $\mathcal{T}$ are transformations with parameters $P \in \mathcal{P}$. Factors of variation are denoted by $F$. \faFire \hspace{0.7pt} indicates trainable parameters, \faSnowflake \hspace{0.7pt} indicates frozen.}
    \label{fig:methods}
\end{figure*}

\section{Related work}\label{sec:related_work}

\subsection{Representation Learning and Conventional Evaluation}

Self-supervised learning (SSL) paradigms have dominated the representation learning field. Their success spans domains like NLP \cite{devlin2019bert, radford2018gpt, raffel2020exploring}, computer vision \cite{he2022masked, assran2023self, chen2020simple}, and speech \cite{hsu2021hubert, chen2022wavlm, baevski2020wav2vec}, in part driven by the flexibility of transformer architectures and the abundance of unlabeled data. They can be broadly categorized into \emph{Joint Embedding} approaches (e.g., SimCLR \cite{chen2020simple}, BYOL \cite{grill2020bootstrap}), which map different views of the same sample to similar locations in latent space while preventing representation collapse, \emph{Generative} approaches (e.g., VAEs \cite{kingma2014auto}, diffusion models \cite{ho2020denoising}), which learn to capture semantic structures by reconstructing or generating data samples, and \emph{Masked Modeling} approaches (e.g., BERT \cite{devlin2019bert}, wav2vec \cite{baevski2020wav2vec}), which predict or reconstruct masked portions of input to capture contextual relationships.

Evaluation frameworks for SSL representations predominantly focus on task performance. Benchmarks in speech \cite{yang2021superb}, audio \cite{turian2022hear}, and vision \cite{zhai2019large} compare models across diverse downstream tasks. They typically increase comprehensiveness by introducing more tasks and datasets, but rarely investigate how models organize information internally. While some benchmarks include robustness evaluations \cite{mirref}, such as testing against domain shifts or noise, the evaluations generally remain task-centric. This downstream-centric approach overlooks critical latent properties like equivariance, invariance, and disentanglement, which can influence generalization, interpretability, and generalizability across tasks. Recent studies \cite{xiao2021what, devillers2023equimod} have explored these properties individually, but a standardized protocol that consolidates these evaluations into a single framework remains absent.

\subsection{Equivariance}

The concept of equivariance has been explored in computer vision through architectural modifications such as group-equivariant convolutions, which embed transformations like color, rotation, and scaling directly into the network design \cite{cohen2016group, lengyel2023color}. Another research direction investigates learned equivariance, where transformations are modeled as learnable operators in the latent space \cite{garridoSelfsupervisedLearningSplit2023, devillers2023equimod}.

In \cite{dangovski2022equivariant}, a contrastive learning approach with a rotation-prediction objective is introduced to conserve rotation-equivariance. The evaluation focuses on transformation-specific tasks, including rotation prediction accuracy and cosine similarity between rotated and non-rotated embeddings. Devillers and Lefort \cite{devillers2023equimod} propose to measure equivariance by comparing displacement between transformed and original embeddings, verifying that transformation information is preserved despite invariance being a viable solution. Further, \cite{garridoSelfsupervisedLearningSplit2023} introduces a method that separates invariant and equivariant embedding components. Evaluation metrics include Mean Reciprocal Rank and Prediction Retrieval Error, assessing transformation retention in the learned representations. Equivariance evaluation typically involves similarity metrics such as cosine similarity and retrieval-based measures. Some approaches quantify equivariance through transformation parameter prediction errors \cite{garridoSelfsupervisedLearningSplit2023}, providing a direct measure of encoded transformation-specific information.

\subsection{Disentanglement}

The formal definition of disentanglement has evolved, with various frameworks attempting to unify previously scattered notions and metrics \cite{sepliarskaiaEvaluatingDisentangledRepresentations2019, do2019theory, eastwood2018framework}. The strict formulation exposed in Subsection \ref{subsubsection: Disentanglement} is often only applicable to toy datasets with predefined factors of variation \cite{sepliarskaiaEvaluatingDisentangledRepresentations2019}. To expand disentanglement to non-trivial examples, further work \cite{eastwood2018framework} decomposes disentanglement into three properties: disentanglement, completeness, and informativeness, respectively evaluating the number of FVs captured by each dimension, the number of dimensions describing each FV, and the information about each FV contained in individual dimensions. Other definitions emphasize conditions where changes in a single latent dimension should affect only one factor of variation and vice versa \cite{sepliarskaiaEvaluatingDisentangledRepresentations2019}. In practice, disentanglement is commonly studied in the context of generative models, where modular, interpretable controls are essential for generating samples with meaningful variations. Factors of variation play a central role in these studies, but disentanglement can also benefit downstream performance, retrieval tasks, and multimodal interaction \cite{eastwood2018framework}.

\section{Methodology}\label{sec:methodology}

An overview of the evaluation methodology is presented in Figure \ref{fig:methods}. We consider pretrained feature extractors $\mathcal{E}$ (See Section \ref{Subsection: Evaluation Details}). Given a data sample in the data space $\mathcal{X}$, $\mathcal{E}$ embeds representations of these data, into an embedding space $\mathcal{Z}$: \begin{equation*}
    \mathcal{E} : \ x \in \mathcal{X} \mapsto z \in \mathcal{Z} \subset \mathbb{R}^d
\end{equation*}
where $d$ is the dimensionality of the latent space and is model-dependent. Sample $x$ is described by factors of variation $F$ in the set of factors of variation (FV) $\mathcal{F}$, which completely describes the data distribution in $\mathcal{X}$. We further identify transformations $\mathcal{T}_{F}$ which modify factor of variation $F$ based on a set of transformation parameters $\mathcal{P}_{\mathcal{T},F} = (p_1,p_2,...p_n)_F$. We propose three new evaluation axes with regard to these identified transformations and FVs: equivariance, invariance, and disentanglement, in addition to informativeness, the standard downstream analysis of the absolute values of FVs, which we extract from datasets specified in Section \ref{Subsection: Evaluation Details}.

The evaluation protocols and metrics we select do not cover the full scope of metrics proposed in the literature to assess equivariance, invariance, and disentanglement. The reasons for this choice are twofold: firstly, we aim to maintain close alignment with the widely understood downstream probing paradigm, making our new axes readily comparable to standard informativeness assessments. Secondly, we want to ensure that performance differences are more clearly attributable to the representations themselves. By using a consistent probe-based methodology across all attributes, we minimize the risk that variations in scores stem from the inherent differences between distinct types of metrics. 

\begin{table*}[t]
    \centering
    \caption{Overview of Image and Speech Models}
    \resizebox{1\textwidth}{!}{
    \begin{tabular}{lcccc}
        \toprule
        \textbf{Model Name} & \textbf{Training Paradigm} & \textbf{Architecture Type} & \textbf{$d$} & \textbf{Training Dataset} \\
        \midrule
        \multicolumn{5}{l}{\textbf{Image}} \\
        \midrule
        ResNet-18/34/50/101 \cite{he2016deep} & Supervised Learning & Convolutional Neural Network & 512/512/2048/2048 & ImageNet \\
        ViT-B-16/B-32/L-16/L-32 \cite{dosovitskiy2021image} & Supervised Learning & Transformer & 768/768/1024/1024 & ImageNet \\
        SimCLR \cite{chen2020simple} & Contrastive Learning & ResNet-50 & 2048 & ImageNet \\
        DINO \cite{caron2021emerging} / DINOv2-s/b/l \cite{oquab2023dinov2} & Contrastive + Distillation & ResNet-50/ViT-S/ViT-B/ViT-L & 2048/384/768/1024 & LVD-142M/ImageNet \\
    
        ViTMAE \cite{he2022masked}& Masked Modeling (MM) & ViT-B-16 & 768 & ImageNet \\
        CLIP \cite{radford2021learningCLIP}& Contrastive Learning & ViT-B-32 & 512 & 400M Image-Text Pairs \\
        IJEPA \cite{assran2023self} & Latent Masked Modeling & ViT-H-14 & 1280 & ImageNet \\
        \midrule
        \multicolumn{5}{l}{\textbf{Speech}} \\
        \midrule
        AudioMAE \cite{huang2022audiomae} & Masked Modeling & ViT-B/16 & 768 & AudioSet \\
        MDuo \cite{niizumi2024m2d}& Contrastive Learning & ViT-B/16 & 3840 & AudioSet \\
        Wav2Vec2 \cite{baevski2020wav2vec}& Latent Masked Modeling & Transformer & 768 & LibriSpeech \\
        HuBERT \cite{hsu2021hubert} & Clustering + MM & Transformer & 768 & LibriSpeech \\
        CLAP \cite{elizalde2023clap}& Contrastive Learning & HT-SAT & 512 & LAION-630k \\
        Whisper \cite{radford2022robust} & Supervised Learning & Transformer & 768 & 680,000 hours of multilingual data \\
        UniSpeech \cite{wang2021unispeech}& Clustering + MM & CNN+Transformer & 768 & 1,350 hours labeled English data \\
        XVector \cite{snyder2018xvector}& Supervised Learning & Time-Delay Neural Network & 512 & Speaker Verification Datasets \\
        \bottomrule
    \end{tabular}
    }
    \label{tab:models_overview}
\end{table*}

\subsection{Equivariance}

Early approaches in previous work measure equivariance to transformation $\mathcal{T}$ as the variation between embeddings $z = \mathcal{E}(x)$ and $z' = \mathcal{E}(\mathcal{T}(x))$. We adopt the interpretation of later work that equivariance represents the ease with which transformation $\mathcal{T}$ can be modeled. To measure equivariance, we propose two tasks, which we aim to keep conceptually similar to downstream probing: \emph{parameter prediction} and \emph{representation prediction}. To perform parameter prediction, we apply transformation $\mathcal{T}_F$ with parameters $P$ sampled from $\mathcal{P}$. We then apply the transformation in the data space and train a shallow Single-Layer Perceptron (SLP) or Multi-Layer Perceptron (MLP) probe on the task of predicting tracked transformation parameters with the concatenation of $z$ and $z'$ as input. Formally, we train $g$ to minimize the parameter equivariance (P-Equivariance) loss:

\begin{equation}
    \mathcal{L}_{E,p} = \mathcal{L}_\mathcal{T}(g(z \oplus z'), P)
\end{equation}

The nature of the loss and the training objective for the projection head are dependent on the parameter, which can be discrete or continuous.\\

The second task is to predict the resulting representation when applying the transformation $\mathcal{T}$ to $x$: R-equivariance or Representation Equivariance. We once again train a shallow probe $g$, this time conditioned on a projection $g_P$ from the parameter space $\mathcal{P}$ to a continuous space $\mathcal{P'}$ : $g_P : P \in \mathcal{P} \mapsto p' \in \mathcal{P}' \subset \mathbb{R}^{d_P}$. The loss is computed between the predicted output of the probe and the true embedding of the transformed sample:
\begin{equation}
    \mathcal{L}_{E,r} = \mathcal{L}_\mathcal{T} (z' , g(z, p'))
\end{equation}

The loss used to train the embedding prediction probe is now adapted to the nature of the representation, meaning that this task can also be applied to discrete representations. We posit that R-equivariance and P-equivariance are not necessarily bijective and represent two interpretations of equivariance, hence our evaluation of both (this is confirmed in Section \ref{subsection: results: equivariance}). The metrics used to evaluate the equivariance of the representations are task-dependent as well, a desirable parallel to downstream probing to foster transferable understanding of mechanisms in downstream probing to this approach. Here we cover data-space transformations, but learned, non-parametric representations to latent space transformations for higher-level concepts can also be evaluated with the non-restrictive formulation of this framework, which we encourage in future work.

\subsection{Disentanglement}

It is difficult to prescribe an ``amount'' of desirable disentanglement for two factors of variation.
Some FVs are not mutually strictly disentangleable in natural data distributions (e.g., dog color and breed). A desirable amount of disentanglement can vary between different FVs and use cases. However, there are many applications that do require a degree of disentanglement between certain factors of variation, such as speech, music, and image generation, semantic retrieval, or style transfer.

To measure disentanglement, we train small probes to predict factors of variation and apply perturbations in the form of controlled transformations (See Table \ref{tab:fv_transformations}) that impact other factors of variation. We then measure the performance loss on the prediction of $F_1$ under the influence of transformation $\mathcal{T}_{F_2}$. Given the limiting factor of the size of probes trained on measuring informativeness, we posit that the change in performance in predicting $F_2$ when $F_1$ changes is indicative of the learned disentanglement of both FVs within the learned representations (as opposed to the probe bottleneck). This approach provides a practical, albeit indirect, measure of disentanglement that is applicable to real-world datasets.

\subsection{Invariance}

To measure invariance to transformation $\mathcal{T}$, we apply the transformation to data samples over a range of parameters (details in Subsection \ref{Subsection: Evaluation Details}) and measure the cosine similarity between the embeddings of the original samples and those of the transformed ones. A higher cosine similarity indicates a more invariant representation---whether that invariance is desirable is, of course, task-dependent.

\section{Experimental setup}\label{sec:experimental_setup}

\subsection{Domains and Factors of Variation}

We elect to constrain our experiments to controllable, simple FVs that are explicitly linked to parametric transformations in the data space as a starting basis for the evaluation we propose. We identify factors of variation from the \emph{Image} and \emph{Speech} domains, which we can easily control to evaluate informativeness, equivariance, invariance, and disentanglement. Table \ref{tab:fv_transformations} summarizes these factors of variation and the associated transformations (as well as which task and evaluation they are used for).

\begin{table}[t]
    \centering
    \caption{Factors of Variation and Transformations.}
    \resizebox{\columnwidth}{!}{%
    \begin{tabular}{l l l l}
        \toprule
        \textbf{FV} & \textbf{Transformation} & \textbf{Value Range} & \textbf{Evaluation} \\
        \midrule
        Hue & HueShift & [-0.5, 0.5] & Inf., Equi., Dis. \\
        Saturation & SaturationShift & [-2, 2] & Inf., Equi., Dis. \\
        Brightness & BrightnessShift & [-2, 2] & Inf., Equi., Dis. \\
        JPEG Compression & Compression & [0, 100] & Inv. \\
        \midrule
        Speech Rate & Time Stretch & [0.5, 2.0] & Inf, Equi., Dis. \\
        Pitch & Pitch Shift & [-12, 12] (semitones) & Equi., Dis. \\
        White Noise & Additive Noise & [-30, 50] dBFs & Equi., Dis.\\
        Reverb & Room Reverb & RT60dB: [0, 3] s & Inv. \\
        \bottomrule
    \end{tabular}%
    }
    \label{tab:fv_transformations}
\end{table}

\subsubsection{Image} We select three well-known factors of variation with influence on color in the image domain: the mean \emph{hue, saturation}, and \emph{brightness}. These FVs control the color gamut of a given image, and as such represent a set of toy FVs to experiment with. Associated transformations are Hue Shift, Brightness Shift, and Saturation Shift (implementations are taken from the \emph{torchvision} transformation library). The value for HueShift is sampled uniformly between -0.5 and 0.5. Both Saturation Shift and Brightness Shift have minimum and maximum values of -2 and 2. We use an additional transformation for Invariance evaluation, JPEG Compression, with a quality factor sampled uniformly between [0,100].
\subsubsection{Speech} We identify the average speech rate in words per second (wps) as a controllable factor of variation in the speech domain. We obtain the ground truth average speech rate by dividing the number of words in the transcripts of individual clips by the duration of the clip. The transformation that can be used to manipulate SR directly is time stretching (TS), the rate for which is randomly sampled between 0.5 (half speed) and 2.0 (double speed). Other transformations for speech include Pitch Shifting (continuous between -12 and 12 semitones), and additive white noise (between -30 and 50 dB) for disentanglement and equivariance. For invariance, we add a reverb transformation. We use the average room parameters from the audiomentations library \cite{audiomentations} and uniformly sample a target RT60dB between 0 and 3 seconds.

With Librispeech \cite{librispeech}, we used the 100-hour clean subset for probe training, the clean development set for validation, and the clean test set for evaluation. With ImageNet-1k-v2 \cite{imagenet}, we used 10\% of the training set for probe training, 1\% of the training set for validation, and the official validation set for evaluation.

\subsection{Evaluation details}\label{Subsection: Evaluation Details}

\subsubsection{Probe training} We extract representations using the feature extractors covered in Table \ref{tab:models_overview}. Probes are trained with an early stopping mechanism conditioned on validation loss, with the Adam optimizer (learning rate $10^{-3}$, weight decay $10^{-3}$) with batch sizes of 32. Probes are either ReLU-activated MLPs or Single-Layer Perceptrons (SLPs). MLPs have hidden units $[512, 256]$ for Informativeness and P-Equivariance, and $[512, 512]$ for R-Equivariance. For R-equivariance, representations are $L_2$-normalized, and the conditioning $\mathcal{T}$ parameters are projected through a single linear layer into a 32-dimensional space before concatenation $z$ as input to the model.

\subsubsection{Metrics} For informativeness, we measure the Mean Squared Error (MSE) between the predicted and ground truth FV. This is specific to our continuous FVs, but is also extendable to classification metrics for discrete FVs. We measure P-Equivariance with MSE between predicted and ground truth parameters. R-Equivariance is evaluated by measuring the MSE and the cosine similarity between predicted embeddings and ground truth modified embeddings after transformation $\mathcal{T}$. Similarly, we use cosine similarity to measure invariance between the transformed and clean embeddings. Lower RMSE indicates higher performance in the Informativeness and P-Equivariance tasks. Higher cosine similarity indicates higher performance in the R-Equivariance and Invariance tasks.

\begin{figure}[t]
    \centering
    \includegraphics[width=0.9\linewidth]{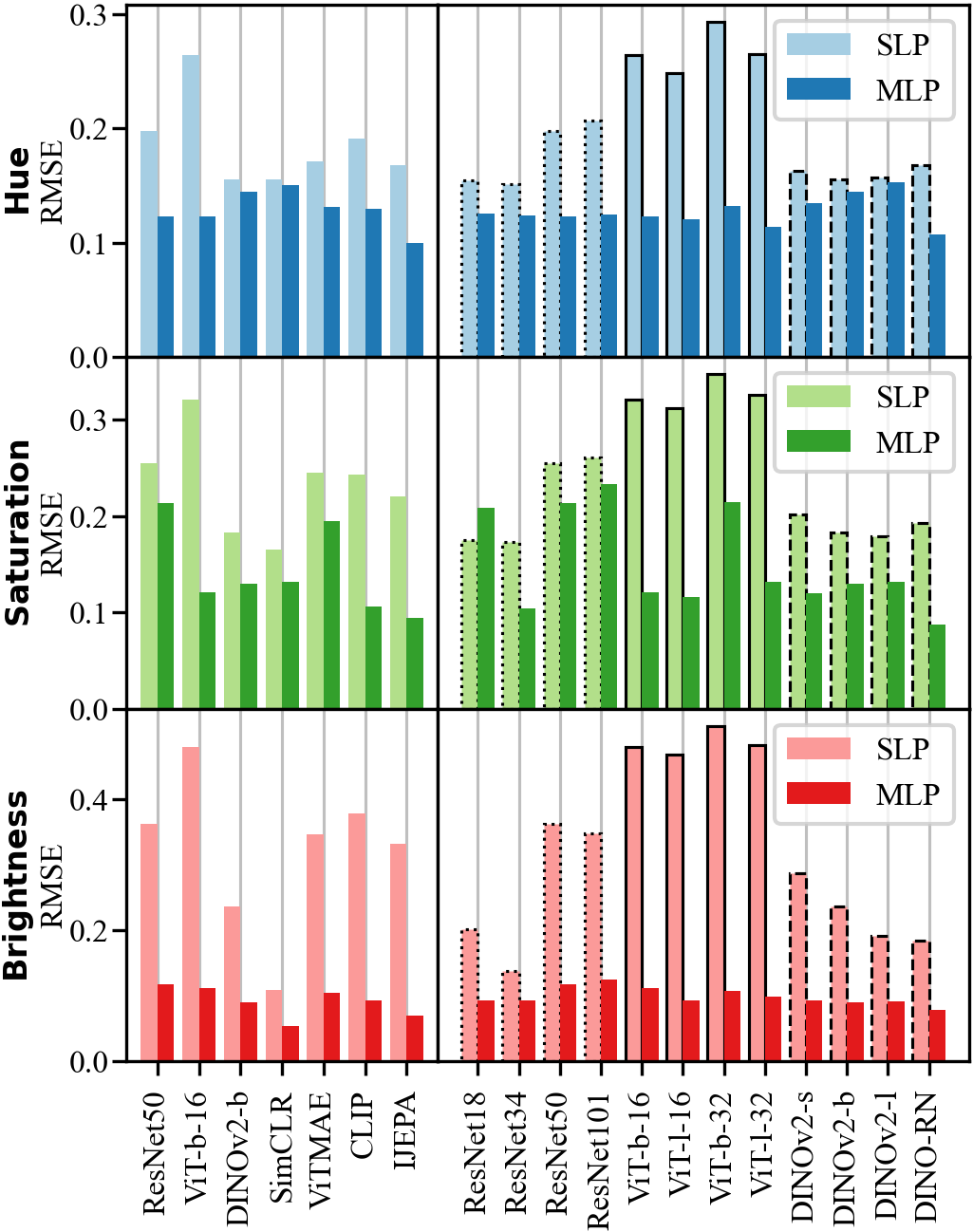}
    \caption{\textbf{Informativeness} on ImageNet for Hue, Saturation, And Value (Brightness) - \emph{left} plot is a comparison of different feature extractors and \emph{right} shows scale ablation.}
    \label{fig:IN_INFO_DOWN_all}
\end{figure}

\section{Results}\label{sec:results}

\subsection{Informativeness}\label{subsection: Informativeness}

\begin{figure}[h]
    \centering
    \includegraphics[width=0.79\linewidth]{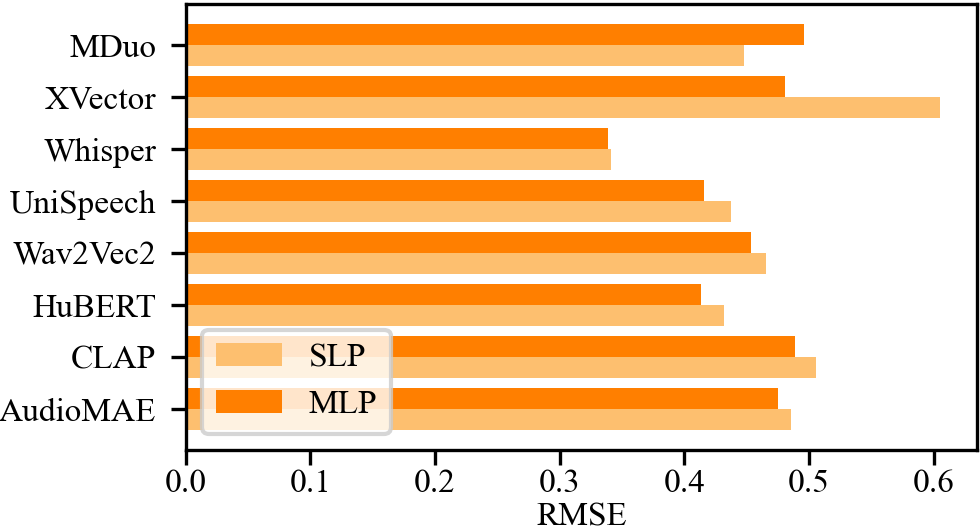}
    \centering
    \caption{Speech rate \textbf{Informativeness} on Librispeech. Speech rate is the only non-normalized (no upper bound) FV, with a mean of $\approx$2.5wps.}
    \label{fig:LS_INFO_DOWN}
\end{figure}

Figures \ref{fig:IN_INFO_DOWN_all}, \ref{fig:LS_INFO_DOWN} present informativeness results on ImageNet and LibriSpeech for various models. Figure \ref{fig:IN_INFO_DOWN_all} includes an ablation study comparing base and larger model variants. Metrics are reported in RMSE for better interpretability.

\begin{figure}[t]
    \centering
    \includegraphics[width=0.9\linewidth]{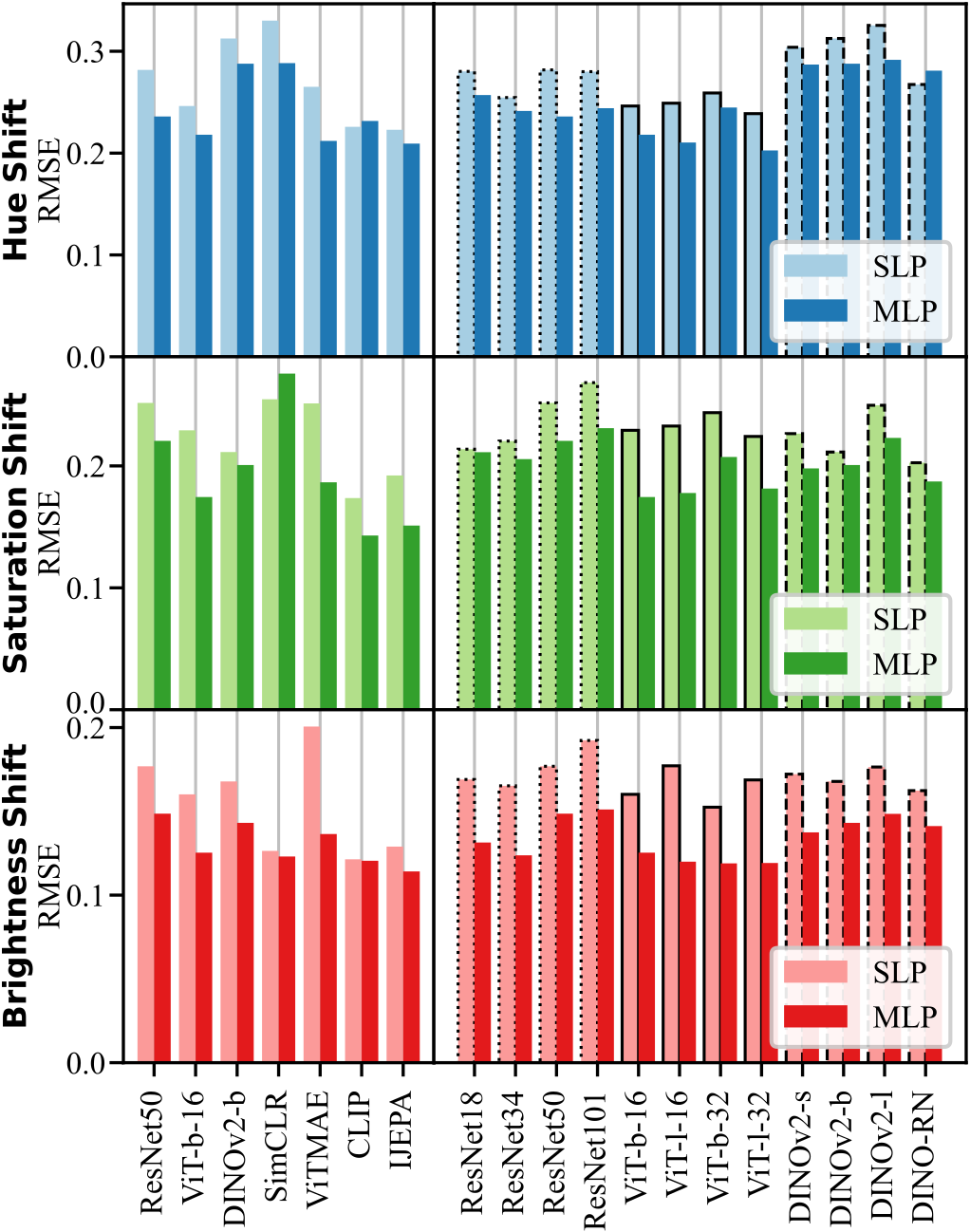}
    \caption{\textbf{Parameter Equivariance} on ImageNet under Hue Shift, Saturation Shift, and Brightness Shift - \emph{left} plot is a comparison of different feature extractors and \emph{right} shows a scale ablation.}
    \label{fig:IN_EQUI_PARA_all}
\end{figure}

\subsubsection{Image} MLPs consistently outperform SLPs across Hue, Saturation, and Brightness FV estimation, though overall trends remain similar. Supervised ViT exhibits the largest differences (13 points for hue, 20 for saturation, 36 for brightness) between SLP and MLP, suggesting color information is encoded nonlinearly rather than following architectural patterns. SimCLR displays symptoms of learned invariance, performing best in brightness estimation but underperforming in hue and saturation (both are used as augmentations during training). Across models, hue estimation shows lower error compared to saturation and brightness. Ablation studies show inconsistent relationships between scale and FV informativeness. ResNet101 generally underperforms its smaller variants across all FVs, particularly when using an SLP, suggesting larger models may encode certain factors less accessibly. In contrast, DINO exhibits increased linear FV availability with larger models, with the DINOv1 ResNet variant outperforming the ViT variants when used with an MLP.

\subsubsection{Speech} Speech rate informativeness closely tracks model scale. Whisper performs best, followed by HuBERT and UniSpeech. XVector, despite being the smallest model, competes with larger models for MLP probing but underperforms in linear evaluation. Spectrogram-based approaches trail tokenized models like Whisper and HuBERT, likely due to the latter's semantic focus. Surprisingly, MDuo, which demonstrates strong performance in general audio tasks \cite{niizumi2024m2d}, ranks among the weakest models for MLP evaluation.

\begin{figure}[t]
    \centering
    \includegraphics[width=0.90\linewidth]{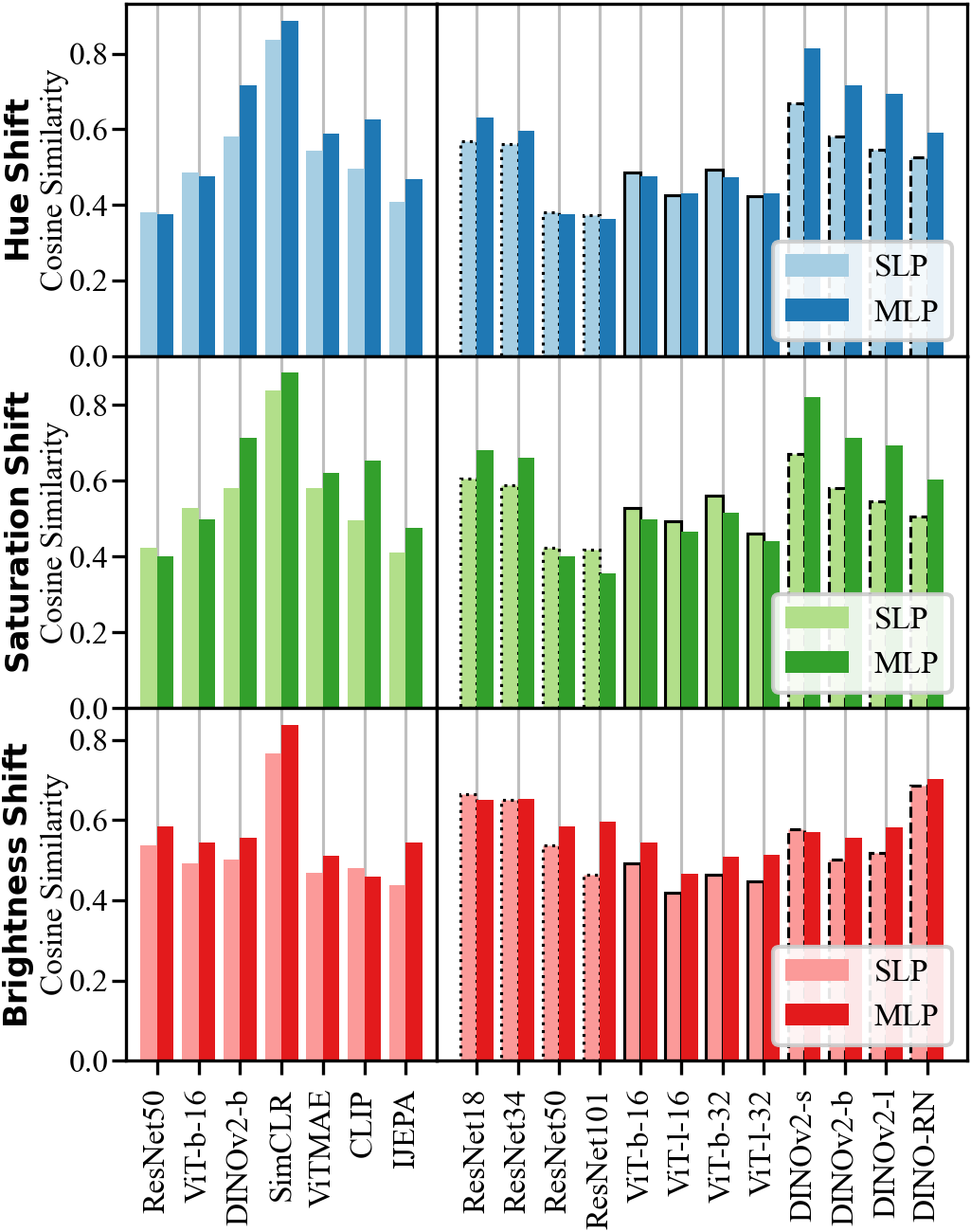}
    \caption{\textbf{Representation Equivariance} on ImageNet under Hue Shift, Saturation Shift, and Brightness Shift - \emph{left} plot is a comparison of different feature extractors and \emph{right} shows a scale ablation.}
    \label{fig:IN_EQUI_FEAT_all}
\end{figure}

\subsection{Equivariance}\label{subsection: results: equivariance}

\begin{figure}
    \centering
    \includegraphics[width=0.9\linewidth]{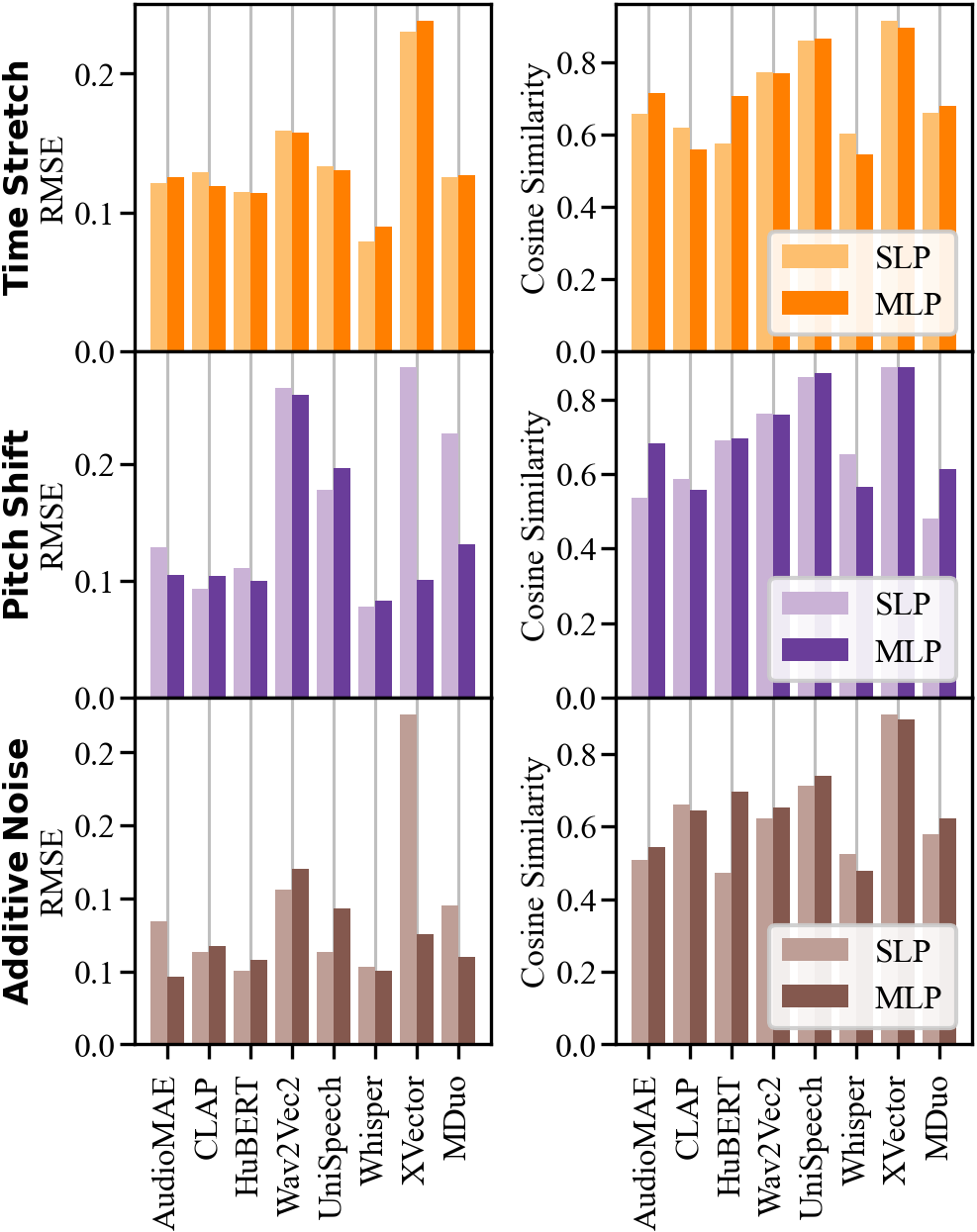}
    \caption{\textbf{P-Equivariance} \emph{(left)} and \textbf{R-Equivariance} \emph{(right)} on Librispeech under Pitch Shift, Time Stretch, and Additive White Noise.}
    \label{fig:LS_EQUI}
\end{figure}

\begin{figure}[h]
    \centering
    \includegraphics[width=0.84\linewidth]{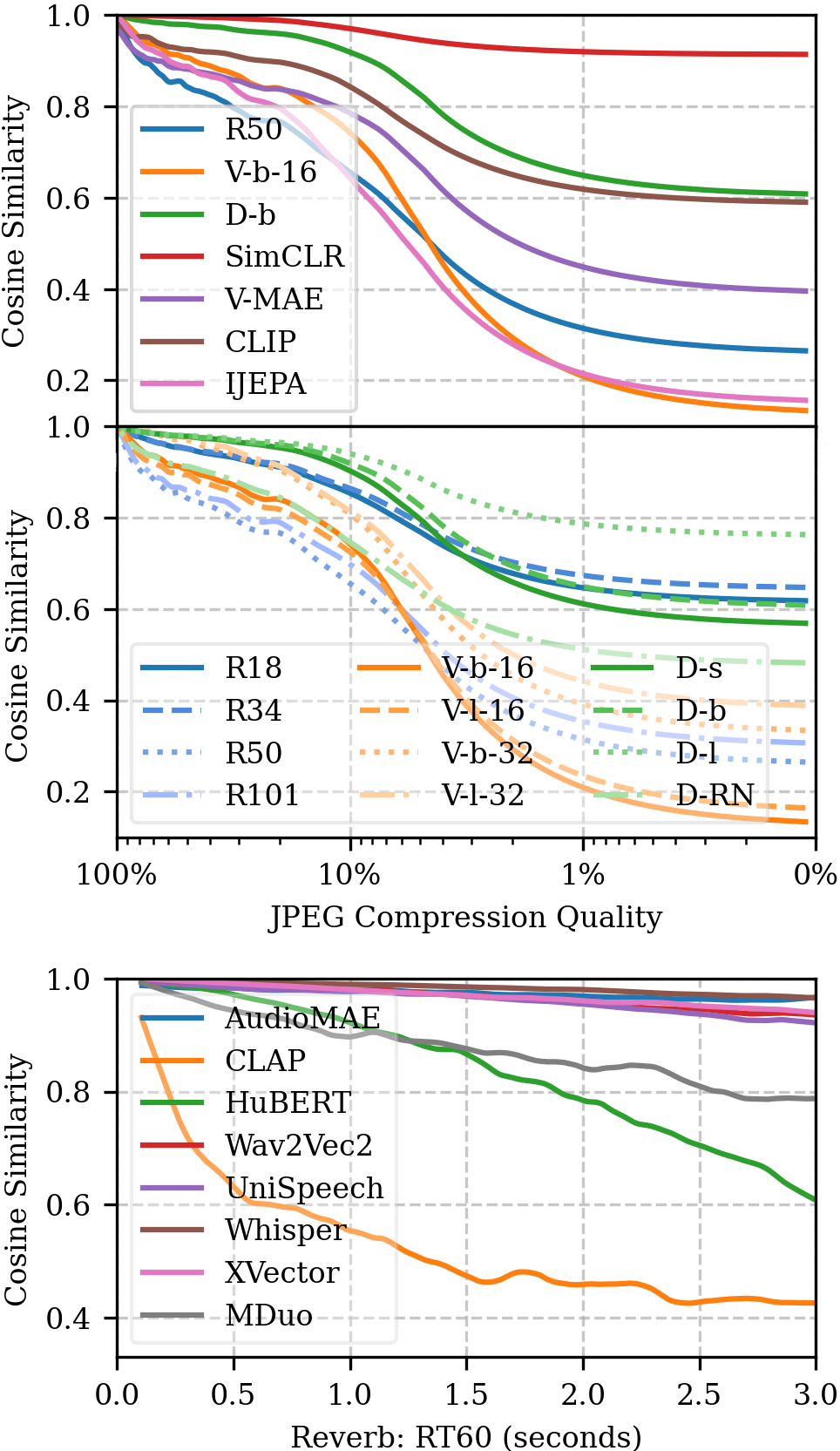}
    \caption{\textbf{Invariance} to Reverb (RT60) on LibriSpeech and \textbf{Invariance} to JPEG compression (\% quality) on ImageNet. Invariance is measured with cosine similarity between transformed and clean representations. Model abbreviations: R is ResNet, V is ViT, D is DINO.}
    \label{fig:INVA}
\end{figure}

\begin{figure*}[t]
    \centering
    \includegraphics[width=1\linewidth]{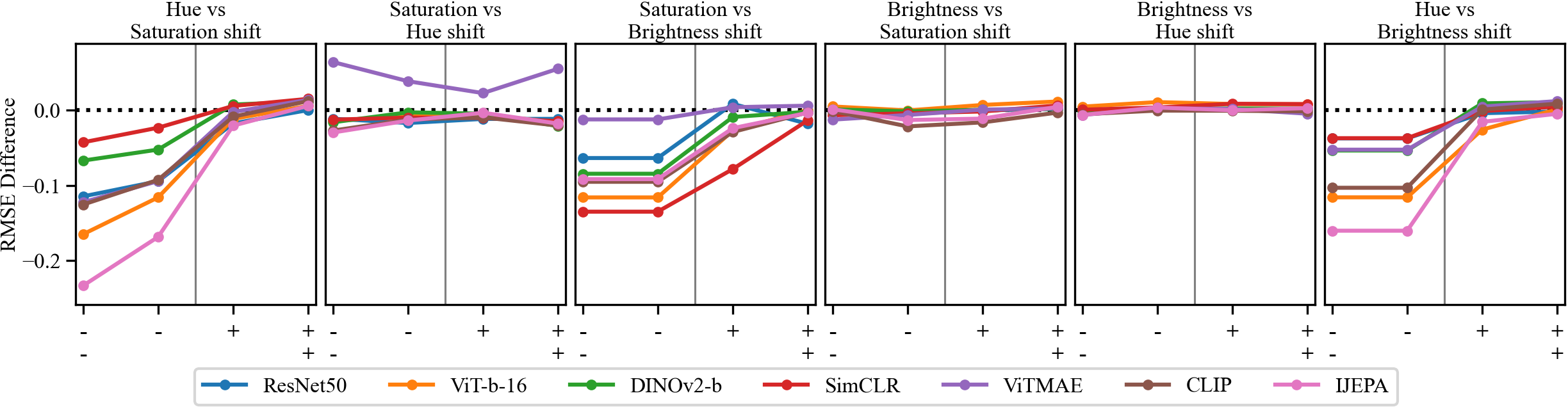}
    \caption{\textbf{Disentanglement} on ImageNet with MLP probes. The y-axis indicates the RMSE difference when predicting FV X after applying a transformation to FV Y (X vs Y), with values closer to zero reflecting better disentanglement. The x-axis represents transformation parameter ranges. \textbf{-- --} : [-100\%, -50\%], \textbf{--} : [-50\%, 0\%], \textbf{+} : [0\%, 50\%], \textbf{+ +} : [50\%, 100\%] of the ranges described in Table \ref{tab:fv_transformations}.}
    \label{fig:DISE}
\end{figure*}

We report P- and R-Equivariance as well as scale ablations in Figures \ref{fig:IN_EQUI_PARA_all} and \ref{fig:IN_EQUI_FEAT_all} for image and \ref{fig:LS_EQUI} for speech.

\subsubsection{Image} We find a much smaller gap in performance for predicting transformation parameters than predicting absolute factors of variation between MLPs and SLPs across the board. Notably, supervised models such as ResNet and ViT are not among the worst performers for P-Equivariance, compared to informativeness. DINO performs notably poorly on Hue shift for P-Equivariance, yet remains competitive on other FVs. SimCLR, another contrastive model trained with random color distortion, also performs poorly on estimating Hue Shift parameters and Saturation Shift parameters as compared to informativeness performance, yet remains competitive on brightness. 

The impact of scale varies by architecture, with ResNet and DINO showing declining P-Equivariance with increased scale, while ViT exhibits no consistent pattern except for brightness, where ViT-L performs worse regardless of patch size.

R-Equivariance seems to be largely dependent on architecture and scale: Contrastive models outperform other models, hinting at a negative correlation between Informativeness and R-Equivariance. Larger scale generally seems to lead to worse R-Equivariance performance (we hypothesize this is due to a more complex latent space landscape), with the exception of DINO with brightness shift.

\subsubsection{Speech} Speech models display varied P-Equivariance behaviors. Whisper leads across all FVs, while XVector performs poorly on pitch shifting and additive noise with an SLP but well with an MLP. Wav2Vec lacks P-Equivariance for pitch shifting but performs well on time-stretching and noise. Notably, SLP probes outperform MLPs for several models (AudioMAE, MDuo, XVector) in white noise and pitch-shifting tasks. This pattern diverges from FV informativeness (Section \ref{subsection: Informativeness}), suggesting structural representation differences despite similar downstream performance. An interesting finding for R-Equivariance is that it seems to be negatively correlated to informativeness, meaning that representations with better speech rate informativeness performance seem to perform worse on understanding how transformations operate in the latent space. This holds particularly true for Time Stretch, which directly modifies the speech rate FV. One possible topological interpretation of this pattern is that latent spaces with strong clustering by FVs may facilitate better factor prediction at the cost of smoothness, which could impact R-Equivariance performance. We encourage future work to verify this hypothesis. Notably, smaller (and less complex) models seem to perform better on R-Equivariance across the board.

\subsection{Invariance}

Invariance results for JPEG compression and Reverb RT60 are shown in Figure \ref{fig:INVA}, as well as an ablation on model scale. Supervised and masked models show lower invariance to JPEG compression, aligning with contrastive learning’s emphasis on transformation invariance. SimCLR’s exposure to blur and color distortions may enhance its robustness despite no direct JPEG training. Scale significantly affects compression invariance: larger ResNet models perform worse, whereas DINO models become more robust with scale. ViT’s invariance increases with larger patches, possibly due to reduced spatial detail. Notably, these invariance patterns contrast with those of equivariance and informativeness, where architecture and scale had less influence.

Speech models also display distinct invariance patterns. AudioMAE, XVector, Wav2Vec, and Whisper exhibit strong robustness to reverb, while HuBERT, MDuo, and CLAP are more susceptible. Mechanistic differences between these behaviors remain unclear. AudioMAE and MDuo both use masked modeling, but AudioMAE focuses on fine-grained reconstruction, while MDuo prioritizes high-level feature retention, potentially leading to reverb sensitivity. CLAP's poor reverb robustness could stem from its training exposure to descriptors referencing reverb.

\subsection{Disentanglement}

Disentanglement results are shown in Figure \ref{fig:DISE}. Model behavior in disentanglement diverges from other axes. DINO consistently achieves strong disentanglement across HSV factors, aligning with its informativeness performance. SimCLR also performs well in disentangling hue and saturation but struggles with saturation when brightness changes, though it successfully disentangles brightness from saturation under saturation changes. ViT and IJEPA perform poorly, exhibiting large prediction changes across most factor pairs (e.g., Hue versus Saturation, Brightness versus Hue, and vice-versa).

Interestingly, disentanglement appears non-bijective: SimCLR poorly disentangles Saturation from Brightness when brightness shift is applied, but disentangles Brightness from Saturation shifts effectively. Some models align with perceptual intuitions; for example, supervised ViT and SimCLR show hue detection difficulty with decreasing saturation, while others like IJEPA and MAE deviate from these perceptual patterns.

\section{Discussion and future work}\label{sec:discussion}

Our evaluation across the four proposed axes demonstrates that models with similar downstream performance in speech and image domains exhibit substantially different behaviors when analyzed through our framework. Even for simple, parametric factors of variation with clear data-space transformations, model behaviors vary unpredictably across axes. This highlights fundamental representation differences that downstream probing alone cannot capture. These differences can guide the selection of representations for specific applications and open avenues for improving trade-offs between informativeness, equivariance, invariance, and disentanglement.

This study has limitations. To maintain conceptual consistency with existing evaluation approaches, we focused on a limited set of factors, metrics, and axes, restricting ourselves to parametric transformations with clear factor correspondences. While these transformations remain simplistic, they provide initial evidence of structural representation differences across models. Our framework is designed to be factor-agnostic, encouraging researchers to extend it with new transformations and performance criteria tailored to their needs.

Future work should explore more complex factors of variation, including non-parametric transformations and higher-level semantic shifts. Extending these axes with new metrics and applying them to broader domains will deepen our understanding of representation structure and behavior. We anticipate that \texttt{synesis} can serve as a foundation for more comprehensive representation evaluation frameworks in diverse applications.

\section{Conclusion}\label{sec:conclusion}

This work has introduced a standardized framework for representation evaluation beyond conventional downstream probing. By introducing the axes of informativeness, equivariance, invariance, and disentanglement, we offer a formalized approach to analyze representation properties that are overlooked in traditional evaluation methods. Through controlled, parameterized transformations and comprehensive experiments across image and speech domains, we demonstrate that representations with similar downstream performance can exhibit significant differences in these structural attributes.

Our findings highlight the necessity of evaluating representations in terms of their internal structure rather than relying solely on task performance metrics. By isolating these core properties from model-specific factors, we provide insights into the underlying mechanisms driving representation behavior. We believe this framework serves as a foundation for more nuanced evaluation protocols and encourages the research community to extend these methods to additional domains, factors of variation, and applications in representation learning.

\clearpage

\bibliographystyle{ieeetr} %
\bibliography{refs}

\end{document}